# POPPINS : A Population-Based Digital Spiking Neuromorphic Processor with Integer Quadratic Integrate-and-Fire Neurons

Zuo-Wei Yeh[1], Chia-Hua Hsu[1], Alexander White[2], Chen-Fu Yeh[2], Wen-Chieh Wu[2], Cheng-Te Wang[2], Chung-Chuan Lo[2], Kea-Tiong Tang[1]

[1] Department of Electrical Engineering, National Tsing Hua University, Hsinchu, Taiwan
[2] Institute of Systems Neuroscience, National Tsing Hua University, Hsinchu, Taiwan

*Abstract*—The inner operations of the human brain as a biological processing system remain largely a mystery. Inspired by the function of the human brain and based on the analysis of simple neural network systems in other species, such as *Drosophila*, neuromorphic computing systems have attracted considerable interest. In cellular-level connectomics research, we can identify the characteristics of biological neural network, called population, which constitute not only recurrent fully-connection in network, also an external-stimulus and self-connection in each neuron. Relying on low data bandwidth of spike transmission in network and input data, Spiking Neural Networks exhibit low-latency and low-power design. In this study, we proposed a configurable population-based digital spiking neuromorphic processor in 180nm process technology with two configurable hierarchy populations. Also, these neurons in the processor can be configured as novel models, integer quadratic integrate-and-fire neuron models, which contain an unsigned 8-bit membrane potential value. The processor can implement intelligent decision making for avoidance in real-time. Moreover, the proposed approach enables the developments of biomimetic neuromorphic system and various low-power, and low-latency inference processing applications.

*Index Terms*—CMOS digital integrated circuits, integer quadratic integrate-and-fire (I-QIF) neuron, low-power design, low-latency inference processor, neuromorphic engineering, population-based spiking neural network.

## I. Introduction

There is growing interest in neuromorphic computing from the inspiration of the human brain, which not only contains one hundred billion neurons and one quadrillion synapses but also handles all-inclusive events and makes precise decision in varying environments with an average power consumption of only 20 Watts [1]. However, because of the complexity of the human brain, biological research is started to focus on other species with simpler brain, such as insects, to enhance the performance of algorithm and design biomimetic system.

Although a *Drosophila* brain is estimated to have only 100,000 to 150,000 neurons, which is considerably much lower than the total neurons in a human brain. Nevertheless, *Drosophila* individuals perform goal-directed navigation and execute precise flight. Recent studies [2–5] on visual neural network systems of *Drosophila* have demonstrated a connectome between ellipsoid body (EB), which is a central complex substructure that shows coupled symmetric and asymmetric circumferentially laminated ring neurons (R-ring and P-ring neurons), and protocerebral bridge (PB), which is a visual information processing in the *Drosophila* brain. Thus, the EB–PB circuitry [6] was constructed with spiking neural networks (SNNs) to realize the functions for prediction and avoidance. Biomimetic SNN-based systems with smaller network (~10 to ~100 neurons) and lower data-transmission bandwidth have a great potential in low-power and low-latency applications.

However, despite lots of research on biological neural networks in visual systems, we still face some difficulties in biomimetic system. First, in biology, neurons are clustered in the form of a "population", which contains external stimulus path, recurrent fully-connection, and self-connection for every neuron in population and is displayed in Fig. 1 [7–8]. The large number of synaptic connections may result in huge efforts for signal processing and calculation. Second, connectome and network function are yet to be fully investigated and are not perfect-defined. Even in the aforementioned EB–PB circuitry, there are some unknown block boxes in visual system, such as lamina monopolar cells and pre-processing visual image (e.g optical flow, etc.) before sending to decision making network.

Therefore, we proposed a configurable population-based digital spiking neuromorphic processor (POPPINS), which is a population-based SNN with an integer quadratic integrate-and-fire (I-QIF) neuron, and postsynaptic decay design to develop biomimetic networks and implementation in any kind of SNN.

## II. Integer Quadratic Integrate-and-fire Neuron

The quadratic integrate-and-fire (QIF) neurons [9] are typically used in recent research designs and has several transformed types such as adaptive QIF neurons (e.g Izhikevich neuron [10]), exponential QIF (Exp-QIF) neurons [11], etc. Most of these require high resolution and quadratic nonlinear multiplication in partial differential function (PDE), as displayed in Fig. 2, which result in large area and high power consumption. For a hardware-friendly equation design, the membrane potential and PDE linearity are necessary to be quantized into low resolution and arithmetic unit. Therefore, the PDE of integer quadratic integrate-and-fire (I-QIF) neuron is designed with an unsigned integer (8-bit) membrane potential $V_m$ and linear conditional equation as following:

$$\Delta V_m = \begin{cases} a(V_r - V_m[t-1]) + I[t], & when\ V_m[t-1] < V_{pde,th} \\ b(V_m[t-1] - V_t) + I[t], & when\ V_m[t-1] \geq V_{pde,th} \end{cases} \quad (1)$$

$$V_m[t] \rightarrow V_{reset}, when\ V_m[t-1] + \Delta V_m > V_{max} \quad (2)$$

where $V_r$, $V_t$, and $V_{reset}$ denote the resting potential, threshold potential, and reset potential, respectively. Here $a, b$ are the neuron parameters with unsigned 3-bit, of which value is from 0 to $\frac{7}{8}$. The condition point $V_{pde,th}$, $\frac{aV_r + bV_t}{a+b}$, for the PDE function is defined by setting $a, b, V_t$, and $V_r$. The next $V_m[t]$ is determined by $V_m[t-1] + \Delta V_m$ or $V_{reset}$, and the neuron exports spike when the accumulation is larger than the maximum value $V_{max}$ in membrane potential, that is 255 in this unsigned 8-bit membrane potential design.

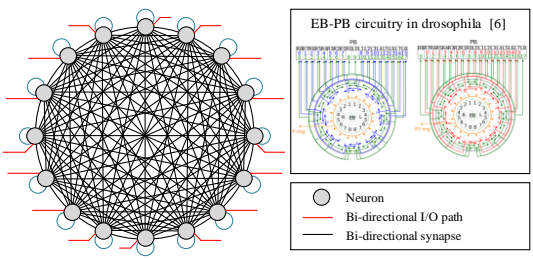

Fig. 1. Population network architecture

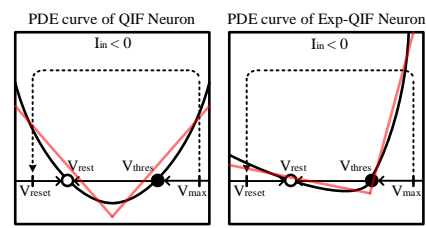

Fig. 2. PDE curve of QIF and Exp-QIF neuron. The red-line is the simplified linear equation with separate point.

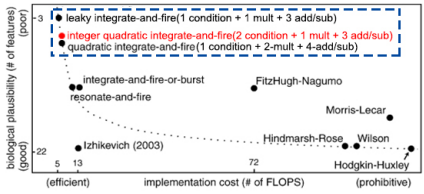

Fig. 3. Comparison of neuron models in terms of biological plausibility and implementation cost

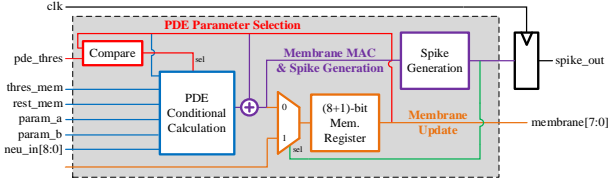

Fig. 4. I-QIF Neuron logic

As displayed in Fig. 3 [12], the I-QIF neuron is more cost-effective (6 FLOPS) than QIF neuron to reach a hardware-friendly design with less multiplication arithmetic unit. Moreover, compared to leaky integrate-and-fire (LIF) neuron, the I-QIF neuron presents more spike behavior than LIF neuron. The neuron logic is designed with 8-bit register and 1-bit overflow in Fig. 4.

## III. PROCESSOR ARCHITECTURE DESIGN

A block diagram of POPPINS is displayed in Fig. 5(a). The overall architecture comprises I/O device (input/output stream buffer), hierarchy-population scheduler, top controller, and two neuromorphic processing units (NPUs), which contain 32 neurons and 128 neurons separately. In addition, 1 neuron is designed for the global excitatory or inhibitory path in each NPU.

### A. Top controller and scheduler

The proposed processor can calculate the synaptic accumulation, decay, and PDE calculation of each neuron continuously, which is the same process in software operation. One time-step is defined as the operation sequence in Fig. 5(c). Therefore, multiple clock cycles are required to compute in hardware. In the time-step loop, the external stimulus, which is from outside of the chip with the address and input-data value, is the first process of synaptic accumulation. Then, the network inter-spike accumulation in population and synaptic decay are checked automatically. At last, the neuron PDE is calculated in the two NPUs simultaneously and the spike events are obtained in the current time-step at the output.

Also, because the neural networks in biology are not formed of one huge population but several small populations, there is hierarchical synaptic connection among NPUs with uni-directional paths. Parallel computing in each population and the design of hierarchy between NPUs with uni-directional synaptic connection reduce the total amount of operations, which reach maximum 25% operation speedup when two NPU enable the same number of neurons. Therefore, the spike events from first- to second-hierarchy population in the previous time-step are stored in the hierarchy-population scheduler and scheduled in the next time-step.

### B. NPU with synapyic weight SRAM

As displayed in Fig. 5(b), the NPU architecture is comprises I/O buffers (a synaptic weight buffer, an external input buffer, and a spike stream buffer), a population controller, a post-synaptic core, and I-QIF neuron cluster. In the NPU, the population local controller can enable only $2^n$ neurons, where n is positive integer, and the related synaptic accumulation and decay calculation.

With the 8-bit membrane potential for accumulation in the I-QIF neuron, the synaptic weight is designed as a signed 4-bit signal (value range from -8 to +7). Each spike event triggers total neuron number of weights (in one NPU) for post-synaptic accumulation. The bandwidth of SRAM is designed to read 8 weights (32-bit) in one clock cycle, which is smaller than the bandwidth of the related weight in one spike event; this not only avoids unnecessary power consumption to load weight with many zeros in one address, but also increases network flexibility in load weights.

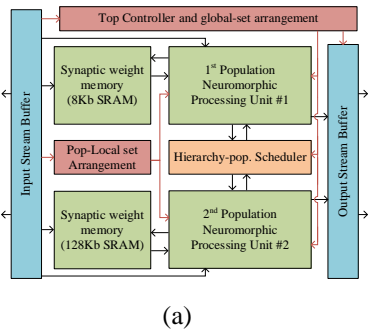
(a)

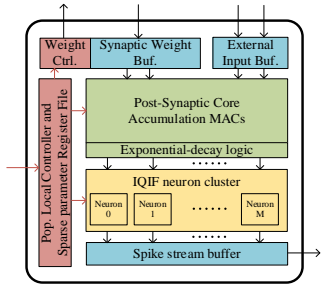
(b)

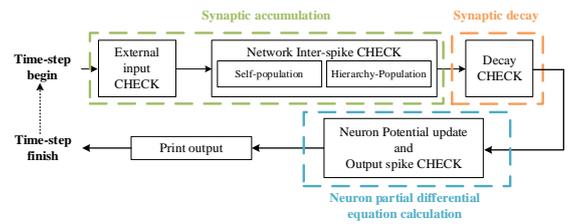
(c)

Fig. 5. (a) Overall block diagram of POPPIINS. (b) NPU architecture (c) Processor strategy of operation sequence to calculate PDE in one time-step loop

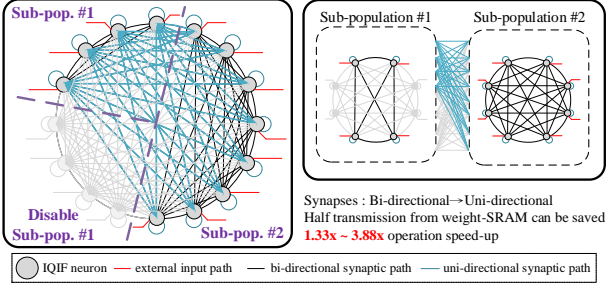

Fig. 6. Illustration of configurable half-hierarchy-chopped population in network view

First, as displayed in Fig. 6, the population in one NPU can be chopped into two small hierarchical sub-populations and each sub-population can be defined active neuron number separately. The recurrent path inside the population is redefined and rearranged at the address of the synaptic weight SRAM and post-synaptic core. With configurable half-hierarchy-chopped population, the speed of the design can be increased by 3.16 times than that of the design without it. Second, Fig. 7(b) shows the group-sparse weight arrangement design. With M neurons in a NPU, there is $\frac{M}{8 \,(SRAM\ bandwidth)}$ groups of post-synaptic path and the group can be triggered by group-sparse code (*GS_code*). The population controller arranges the weights into related post-synaptic path to enable accumulation.

For synaptic accumulation in the population, the crossbar architecture, as displayed in Fig. 7(a), is implemented for system design. For example, the first NPU has N neurons and second NPU has *M* neurons, post-synaptic core emulates *N×N* and *(N+M)×N* crossbar in each NPU. Every feed-forward and backward spike event can be directly transmitted as a crossbar input to direct address of the related weight line for accumulation in each knot of the crossbar. After that, the crossbar output update the post-synaptic potential. Although, the crossbar can achieve high-speed calculation, it requires a large input/weight transmission bandwidth and increases the silicon area. Therefore, the most crossbar architecture is virtualized to one-bar $(1 \times N)$ parallel accumulation with synaptic weight SRAM and spike decoder which is displayed in Fig. 7(b). In NPU, a spike event needs several clock cycles, which is named MAC-cycle and determined by the group-sparse number (*GS_num*) setting in population controller, to do post-synaptic accumulation. The spike decoder check the spike stream triggered in the previous time-step, spike decoder detect the LSB 2-bit spike stream register at every clock cycle to determine MAC-cycle needed to wait and the weight address to the synaptic SRAM. Finally, the spike decoder performs the circular-shift to read next 2-bit spike stream until all spike stream is checked. Then, the operation counter in the population controller counts up and determines the state of the *decay_en* signal in Fig. 8. As shown in Eq. (3), the reciprocal decay logic is designed to decrease value with the power of two, which is achieved by using shifter and determined by *decay_α*. However, with digitalization in decay logic from Eq. (3) to Eq. (4), the decay fatigue is happened to digital reciprocal decay logic because of value-vanishing in $y[t-1] \gg decay\_\alpha$. Therefore, the equation is modified to Eq. (5) to avoid decay fatigue with the selector logic between shift result ($y[t-1] \gg decay\_\alpha$) and $decay_{min}$ value, which is ±1 determined by signed-bit of *y[t-1]*.

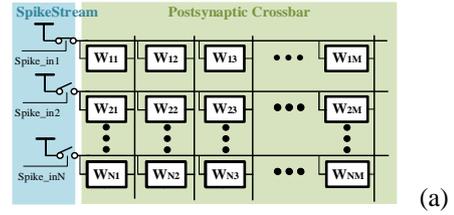

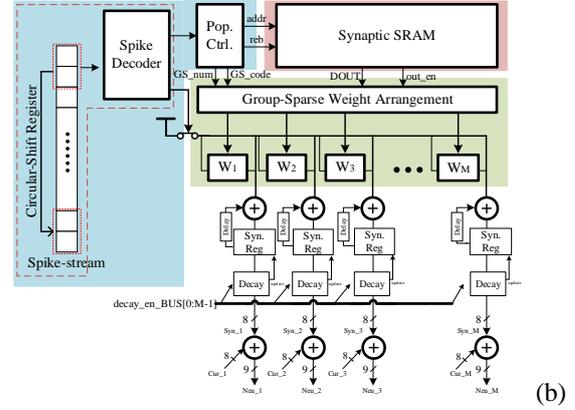

Fig. 7. (a) the ideal system design with post-synaptic crossbar architecture (b) Virtualized crossbar architecture with spike decoder and on-chip SRAM

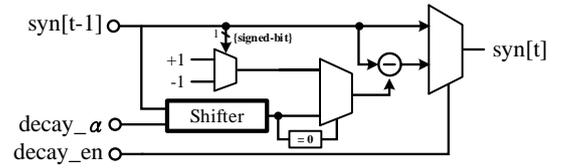

Fig. 8. Reciprocal decay logic

$$y[t] = y[t-1] * (2^{decay\_\alpha} - 1)/2^{decay\_\alpha} \qquad (3)$$

$$y[t] = y[t-1] - y[t-1] \gg decay\_\alpha \qquad (4)$$

$$y[t] = y[t-1] - SEL(y[t-1] \gg decay\_\alpha, decay_{min}) \ (5)$$

In some applications, reciprocal decay logic can be used to approach the exponential decay via the required time of 10% decrease to imitate the real synapse in biology.

IV. SPECIFICATION, MEASUREMENT AND RESULT

POPPINS is fabricated in TSMC 0.18-µm CMOS process. The die photo and technological specification are presented in Fig. 9. We not only measured the behavior model of I-QIF neuron but also the related specification of single neuron area and power consumption. Furthermore, we tested POPPINS by implementing two different models for solving Sudoku puzzle and obstacle avoidance decision-making.

**POPPINS Chip Characteristic**

| | |
|---|---|
| Implementation | Digital |
| Technology | 0.18µm |
| Area (incl. pads) | 8.32 mm² |
| Power Supply | 1V |
| Max. clock Frequency | 100MHz |
| Total SRAM memory | 136 Kb |
| Power Consumption | 107mW |
| # Neurons / # Synapses | 162 / 17.73k |
| Energy Efficiency @ 100MHz<br>33 Neurons @ NPU1+129 Neurons @ NPU2 | 69.13pJ @1V |
| 4x4 Sudoku puzzle speed@ 100MHz<br>64 Neurons @ NPU2 with 50% sparsity | ~13.4K puzzles/sec |
| Decision making speed@100MHz<br>8 Neurons @ NPU1 | 29.4K decisions/sec |

Fig. 9. POPPINS die-photo and specification result

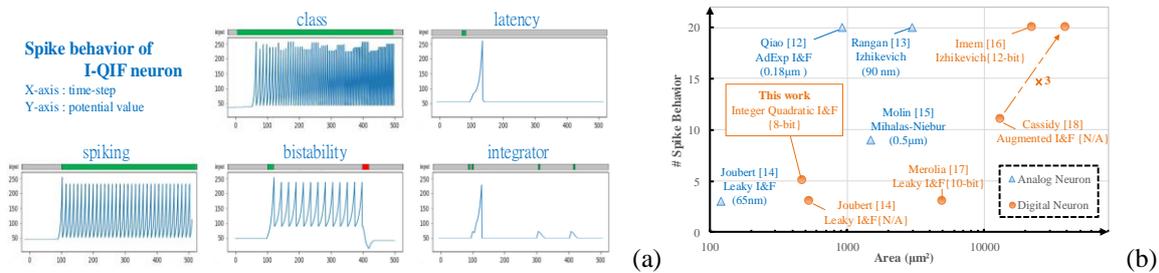

Fig. 10. (a) The measurement for 5 spike behaviors of I-QIF (b) State of the art of neuron implementation about area and spike behaviors which can be outputted from neuron. The area of digital designs are normalized to a 65nm node.

## A. I-QIF Neuron

Fig. 10(a) displays the five spike behaviors of the proposed neuron. Moreover, the I-QIF neuron specification and overview of the current state for both analog and digital neurons [13–19] are presented in Fig. 10(b). Compared with other neurons such as leaky I&F neuron, the I-QIF neuron is more efficient in area normalized to the same process [20], which is positive correlation to power consumption. The detailed specification and performance of the I-QIF neuron are summarized in Table I.

## B. Sudoku puzzle

N×N Sudoku puzzles from 2×2 to 5×5 were solved to verify the proposed POPPINS. A recurrent neural network with total $N^3$ neurons and $N^6$ synapses with excitatory and inhibitory connections were mapped to every digit in the Sudoku puzzle [21]. Fig. 11(a) displays the architecture of network with $4^3$ neurons to solve 4×4 Sudoku puzzle and the excitatory and inhibitory paths within one neuron's output connections in both the same layer and different digit layers. Furthermore, each digit in one position connected only neurons in eight position and with group-sparse setting, the operation speed could be increased by near 2 times. Finally, Fig. 11(b) displays the output spike result of one digit layer in the case of Fig. 11(a).

## C. Decision making for avoidance of biomimetic

Studies have investigated the visual system of *Drosophila* and reconstructed the system [21–23]. In this study, the image sensor replaced retina cells, optical flow replaced the cells from Lamina to Lobula cells, and matrix dot operation replaced Lobula plates. For obstacle avoidance decision-making, eight motions were performed (can expand the motions if necessary). As displayed in Fig. 12, the same stimulus requires 50 time steps in each decision result for accumulation to export spike stream. Therefore, the decision latency was 1/300 s in an emergency, which resulted in a real-time reaction time of >30 fps. At last, Table I provides specification and measurement benchmark results compared to others SNN processors, including DYNAPs [25], IFAT [26], Truenorth [27], Loihi [28], and MorphIC [29].

## V. CONCLUSION

In this study, POPPINS was designed with a virtualized crossbar to accumulate postsynaptic potential value in 180-nm technology. POPPINS can be used in various networks and applications. To verify the implementation of the algorithm in hardware design, POPPINS was used to perform different tasks (Sudoku puzzle and avoidance) in a time-constrained manner. The proposed processor achieved a high efficient and real-time computation with a supply voltage of 1 V and a maximum clock frequency of 100 MHz.

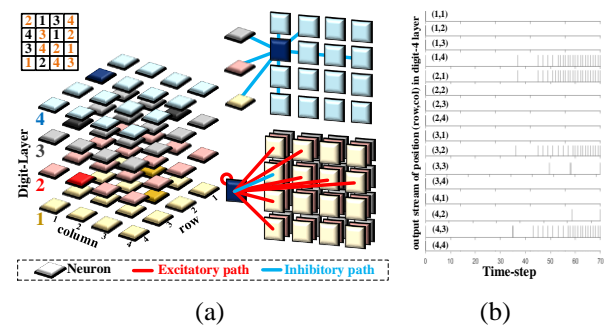

Fig. 11. (a) SNN to solve 4×4 Sudoku puzzle and synaptic connection from a neuron in digit-4 layer. (b) the output waveform in digit-4 layer

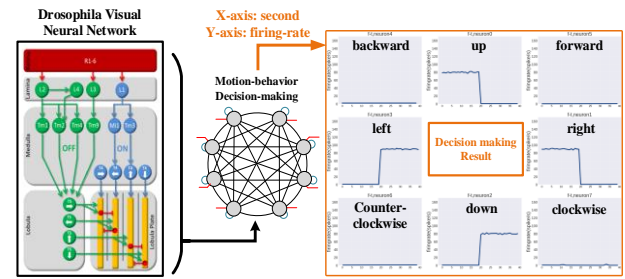

Fig. 12. Movement in eight directions after visual neural network processing or pre-processor with 1MHz clock frequency (~15000 time-steps/s) when a new stimulus imports into the network in half time.

TABLE I. COMPARISION OF THE SNN PROCESSOR CHIPS

| Reference | [25] | [26] | [27] | [28] | [29] | This work |
|---|---|---|---|---|---|---|
| Technology(nm) | 180 | 90 | 28 | 14 | 65 | 180 |
| Implementation | Mixed | Mixed | Digital | Digital | Digital | Digital |
| Area (mm$^2$) | 43.79 | 16 | 389 | 51.8 | 2.86 | 8.32 |
| Syn. density (syn/mm$^2$)* | 2.1k | N/A | 125k | Max. 190k | 738k | 33.6k |
| Neuron. density (neu/mm$^2$)* | 34 | 4k | 494 | Max. 190 | 716 | 259 |
| Neu. Behavior | 20 | 3 | 11 | N/A | N/A | 5 |
| Weight Storage | 12b | off-chip | 1b | 1b~9b | 1b | 4b |
| Decay. Synapse | × | × | × | O | × | O |
| Total Energy efficiency* | 30pJ/ SOP | 22pJ/ SOP | 52pJ/ SOP @0.775V | N/A | 51pJ/ SOP @0.8V | **13.2pJ/ SOP @ 1V** |

* The digital-implemented chip are normalized to 65nm process in area and power domain.

ACKNOWLEDGMENT

This work was supported by the Ministry of Science and Technology, Taiwan, under contract no. MOST 109-2218-E-007-019 and MOST 108-2262-8-007-017. This work was supported by Taiwan Semiconductor Research Institute (TSRI) for chip fabrication.